 \documentclass[tablecaption=bottom,wcp]{jmlr} % W&CP article

\usepackage{booktabs}
\usepackage[load-configurations=version-1]{siunitx} % newer version
\newcommand{\mr}[1]{\mathrm{#1}}

\usepackage{multirow}

\theorembodyfont{\upshape}
\theoremheaderfont{\scshape}
\theorempostheader{:}
\theoremsep{\newline}

\jmlrproceedings{AABI 2020}{3rd Symposium on Advances in Approximate Bayesian Inference, 2020}
\title[Invertible DenseNets]{Invertible DenseNets}

  \author{\Name{Yura {Perugachi-Diaz}} \Email{y.m.perugachidiaz@vu.nl}\\
  \Name{Jakub {M. Tomczak}} \Email{j.m.tomczak@vu.nl}\\
  \Name{Sandjai {Bhulai}} \Email{s.bhulai@vu.nl}\\
  \addr Vrije Universiteit Amsterdam, Netherlands}

\begin{document}

\maketitle

\begin{abstract}
We introduce Invertible Dense Networks (i-DenseNets), a more parameter efficient alternative to Residual Flows. The method relies on an analysis of the Lipschitz continuity of the concatenation in DenseNets, where we enforce the invertibility of the network by satisfying the Lipschitz constraint. Additionally, we extend this method by proposing a learnable concatenation, which not only improves the model performance but also indicates the importance of the concatenated representation. We demonstrate the performance of i-DenseNets and Residual Flows on toy, MNIST, and CIFAR10 data. Both i-DenseNets outperform Residual Flows evaluated in negative log-likelihood, on all considered datasets under an equal parameter budget. 
\end{abstract}

% Keywords may be removed
% \begin{keywords}
%List of keywords
%\end{keywords}

%%%%%%%%%%%%%%%%%%%%%%%%%%%%%%%%%%%%%%%%%%%%%%%%%%%%%%%%%%%%%%%%%%%%%%
%                       Introduction                                %
%%%%%%%%%%%%%%%%%%%%%%%%%%%%%%%%%%%%%%%%%%%%%%%%%%%%%%%%%%%%%%%%%%%%%%
\section{Introduction} \label{sec:introduction}
Neural networks are frequently used in supervised learning tasks such as classification, where models are trained to predict labels. However, they are also used to parameterize generative models that try to estimate the true distribution of the observed data. Generative models can be used to generate realistic-looking images that are hard to separate from real ones, detection of adversarial attacks~\citep{fetaya2019understanding, jacobsen2018excessive}, and for hybrid modeling~\citep{nalisnick2019hybrid} which have the property to both classify and generate.

The generative architecture come in different designs. A common approach to train generative models is using the likelihood objective. One kind of model that also uses this approach are flow-based models.
Flow-based models consist of invertible transformations that allow them to compute the likelihood using the change of variable formula. The main difference that determines an exact computation or approximation of a flow-based model, lies in the design of the transformation layer. The design used to make this layer invertible can consist of the exact computation of the inverse or a numerical technique. For example,~\citep{realnvp} use coupling layers that consist of functions stacked on each other to make the flow invertible. This allows an exact computation while modeling complex convolutional neural networks that do not require the computation of the derivative.

Recently,~\citet{i-resnet} have proposed deep-residual blocks as a transformation layer. The deep-residual networks (ResNets) of~\citep{he2016deep} are known for their successes in supervised learning approaches. In a ResNet block, each input of the block is added to the output, which forms the input for the next block. Since ResNets are not necessarily invertible,~\citet{i-resnet} enforce the Lipschitz constraint in such a manner that the network becomes invertible. Furthermore,~\citet{chen2019residualflows} proposed Residual Flows, an improvement of i-ResNets, that uses an unbiased estimator of the log-determinant, which results in even better performance. 

In supervised learning, an architecture that uses fewer parameters and is even more powerful than the deep-residual network, is the Densely Connected Convolution Network (DenseNet), which was first presented in~\citep{huang2017densely}. The network showed to improve significantly in recognition tasks on benchmark datasets such as CIFAR, SVHN, and ImageNet, by using fewer computations and having fewer parameters than ResNets while performing at a similar level.
In contrary to a ResNet block, a DenseNet layer consists of a concatenation of the input with the output. In this work, we introduce invertible Dense Networks (i-DenseNets), and we show that we can enforce the Lipschitz constraint in a similar manner as in~\citep{i-resnet}. Further, we show that this model can be efficiently trained as a generative model and outperforms Residual Flows under an equal parameter budget. 
%%%%%%%%%%%%%%%%%%%%%%%%%%%%%%%%%%%%%%%%%%%%%%%%%%%%%%%%%%%%%%%%%%%%%%
%                           Background                               %
%%%%%%%%%%%%%%%%%%%%%%%%%%%%%%%%%%%%%%%%%%%%%%%%%%%%%%%%%%%%%%%%%%%%%%
\section{Background}
Let us consider a vector of observable variables $x \in \mathbb{R}^d$ and a vector of latent variables $z \in \mathbb{R}^d$.
We define a bijective function $f: \mathbb{R}^d \rightarrow \mathbb{R}^d$ which maps a latent variable to datapoint $x = f(z)$. If $f$ is invertible, we define its inverse as $F = f^{-1}$. Further, we use the change of variable formula to compute the likelihood of a datapoint $x$ by:
\begin{equation} \label{eq:covf}
    \ln p_X(x) = \ln p_Z(z) + \ln | \det J_F(x) |, 
\end{equation}
where $p_Z(z)$ is a base distribution (e.g., the standard Gaussian) and
$J_F(x)$ is the Jacobian of $F$ at $x$. The change of variable formula allows tractable evaluation of the data and the flows are trained using the maximum likelihood objective.

\citet{i-resnet} construct an invertible ResNet layer which is only constraint in Lipschitz continuity. A ResNet is defined as: $F(x) = x + g(x)$, where $g$ is modeled by a (convolutional) neural network and $F$ represents a ResNet layer which is in general not invertible. However, they construct $g$ in such way to satisfy $\mathrm{Lip(g)} < 1$ by using spectral normalization of~\citep{gouk2018regularisation, miyato2018spectral}, such that:
\begin{equation} \label{eq:spec_norm}
    \mathrm{Lip}(g) < 1, \quad \mathrm{if}\quad ||W_{i}||_2 < 1,
\end{equation}
where $|| \cdot ||_2$ is the $\ell_2$ norm. 
Since the Banach fixed-point theorem holds in this specific case, the ResNet layer $F$ has a unique inverse, even though there does not need to be an analytical closed-form solution. Further, the log-determinant can be estimated using the Hutchinsons trace estimator~\citep{skilling1989eigenvalues, hutchinson1989stochastic}, at a lower cost than to fully compute the trace of the Jacobian. \citet{chen2019residualflows} propose Residual Flows that uses an improved method to estimate the log-determinant with an unbiased estimator.

%%%%%%%%%%%%%%%%%%%%%%%%%%%%%%%%%%%%%%%%%%%%%%%%%%%%%%%%%%%%%%%%%%%%%%
%                       Invertible DenseNets                         %
%%%%%%%%%%%%%%%%%%%%%%%%%%%%%%%%%%%%%%%%%%%%%%%%%%%%%%%%%%%%%%%%%%%%%%
\section{Invertible Dense Networks} \label{sec:invertible_densenet}
We introduce i-DenseNets, an invertible model based on DenseNets parametrizations. To formulate i-DenseNets, we define a DenseBlock as a function $F: \mathbb{R}^d \rightarrow \mathbb{R}^d$ with $F(x) = x + g(x)$, where $g$ consists of Dense Layers $\{h_i\}_{i=1}^n$  that are expressed as:

\begin{equation} \label{eq:dn_layer}
        g(x) = h_{n+1} \circ h_{n} \circ \cdots \circ h_1(x), 
\end{equation}
where $h_{n+1}$ represents a $1 \times 1$ convolution to match the output size of $\mathbb{R}^d$. A layer $h_i$ consist of two parts concatenated to each other. The upper part is a copy of the input signal. The lower part consist of the transformed input, where the transformation is a multiplication of (convolutional) weights $W_i$ with the input signal, followed by a non-linearity $\phi$ having $\mr{Lip}(\phi) \leq 1$, such as ReLU, ELU, LipSwish, or tanh. 
As an example, a Dense Layer  $h_2$ can be composed as follows:
\begin{equation}
    h_1(x) = \begin{bmatrix} 
    x \\ \phi(W_1x)
    \end{bmatrix}
    , \quad
    h_2(h_1(x)) = \begin{bmatrix} 
    h_1(x) \\ \phi(W_{2} h_1(x))
    \end{bmatrix}.
\end{equation}

\subsection{Enforcing Lipschitz constraint} \label{sec:enforcing_lip}
If we enforce the function $g$ to satisfy $\mathrm{Lip}(g) < 1$, the DenseBlock $F$ is invertible and the Banach fixed point theorem holds. As a result, the inverse can be approximated in the same manner as in~\citep{i-resnet}. To satisfy $\mr{Lip}(g) < 1$, we can enforce $\mr{Lip}(h_i)<1$ for all $n$ layers. Therefore, we first need to determine the Lipschitz constant for a Dense Layer $h_i$. We know that a function $f$ is $\mr{K}$-Lipschitz if for all points $v$ and $w$ the following holds (for the full derivation see Appendix~\ref{apd:A_appendix}):
\begin{equation} \label{eq:lip}
    d_Y(f(v), f(w)) \leq \mr{K} d_X(v, w),
\end{equation}
where we assume that the distance metrics $d_X=d_Y=d$ are chosen to be the $\ell_2$-norm. Further, let two functions $f_1$ and $f_2$ be concatenated in $h$:
\begin{equation} \label{eq:dn_functions}
    h_v = 
    \begin{bmatrix}
    f_1(v)\\ f_2(v)
    \end{bmatrix},
    \quad
    h_w =
    \begin{bmatrix}
    f_1(w)\\ f_2(w)
    \end{bmatrix},
\end{equation}
where function $f_1$ is the upper part and $f_2$ is the lower part.
We can now find an analytical form to express a limit on $\mr{K}$ for the Dense Layer in the form of~\equationref{eq:lip}:
\begin{equation}
\begin{split}
    d(h_v, h_w)^2 &= d(f_1(v), f_1(w))^2 + d(f_2(v), f_2(w))^2, \\
    d(h_v, h_w)^2 &\leq (\mr{K}_1^2 +\mr{K}_2^2) d(v, w)^2,
\end{split}
\end{equation}
where we know that the Lipschitz constant of $h$ consist of two parts, namely, $\mr{Lip}(f_1)=\mr{K}_1$ and $\mr{Lip}(f_2)=\mr{K}_2$. Therefore, the Lipschitz constant of layer $h$ can be expressed as:
\begin{equation}
    \mathrm{Lip}(h) = \sqrt[\leftroot{-1}\uproot{1}]{(\mr{K}_1^2 +\mr{K}_2^2)}.
\end{equation}
With spectral normalization of~\equationref{eq:spec_norm}, we know that we can enforce (convolutional) weights $W_i$ to be at most $1$-Lipschitz. Hence, for all $n$ Dense Layers we apply the spectral normalization on the lower part which locally enforces $\mr{Lip}(f_2)=\mr{K}_2 < 1$. Further, since we enforce each layer $h_i$ to be at most $1$-Lipschitz and we start with $h_1$, where $f_1(x)=x$, we know that $\mr{Lip}(f_1)=1$. Therefore, the Lipschitz constant of an entire layer can be at most $\mr{Lip}(h) = \sqrt[\leftroot{-1}\uproot{1}]{1^2+1^2} =\sqrt[\leftroot{-1}\uproot{1}]{2}$, thus dividing by this limit enforces each layer to be at most $1$-Lipschitz.

\subsection{Learnable concatenation}
We have shown that we can enforce an entire Dense Layer to have $\mr{Lip}(h_i) < 1$ by applying a spectral norm on the (convolutional) weights $W_i$ and then divide the layer $h_i$ by $\sqrt{2}$. To optimize and learn the importance of the concatenated representations, we create learnable parameters $\eta_1$ and $\eta_2$ for, respectively, the upper and lower part of each layer $h_i$. Since the upper and lower part of the layer can be at most 1-Lipschitz, multiplication by these factors results in functions that are at most $\eta_1$-Lipschitz and $\eta_2$-Lipschitz. From~\appendixref{apd:A_appendix} we know that the layer is then at most $\sqrt{\eta_1^2 + \eta_2^2}-$Lipschitz. Dividing by this factor results in a bound that is at most $1$-Lipschitz. 
\begin{figure}[htbp]
\floatconts
    {fig:learnable_params}
    {\caption{Range of the possible normalized parameters $\hat{\eta}_1$ and $\hat{\eta}_2$.}}
    {\includegraphics[width=.23\textwidth]{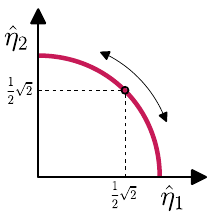}}
\end{figure}

\noindent In practice, we initialize $\eta_1$ and $\eta_2$ at value $1$ and during training use a softplus function to avoid them being negative. The range of the normalized parameters is between $0 \leq \hat{\eta}_{1}, \hat{\eta}_{2} \leq 1$ and can be expressed on the unit circle as is shown in~\figureref{fig:learnable_params}. In the special case where $\eta_1 = \eta_2$, the normalized parameters are $\hat{\eta}_1=\hat{\eta}_2=\frac{1}{2} \sqrt{2}$. This case corresponds to the situation in Section~\ref{sec:enforcing_lip} where the concatenation was not learned. An additional advantage is that the normalized $\hat{\eta}_1$ and $\hat{\eta}_2$ express the importance of the upper and lower signal. For example, when $\hat{\eta}_1 > \hat{\eta}_2$, the input signal is of more importance than the transformed signal.

%%%%%%%%%%%%%%%%%%%%%%%%%%%%%%%%%%%%%%%%%%%%%%%%%%%%%%%%%%%%%%%%%%%%%%
%                           Experiments.                             %
%%%%%%%%%%%%%%%%%%%%%%%%%%%%%%%%%%%%%%%%%%%%%%%%%%%%%%%%%%%%%%%%%%%%%%
\section{Experiments} \label{sec:experiments}
To make a clear comparison between the performance of Residual Flows and i-DenseNets, we train both models on 2-dimensional toy data and on high-dimensional image data, MNIST and CIFAR10. To benchmark the models, we use the architecture of Residual flow~\citep{chen2019residualflows}. 
Since we have a constrained computational budget, we use a smaller architecture of the model and choose number of scales for the toy data and image data set to, respectively, 10 blocks and 4 blocks per 3 scales instead of 100 blocks and 16 blocks per 3 scales. For the other arguments, default settings are used. To compare Residual Flows with i-DenseNets, we utilize an architecture that uses a similar number of parameters for each dataset trained on. A detailed description of this architecture can be found in~\appendixref{apd:B_appendix}. Furthermore, we add the option to learn the parameters of the concatenation. The models trained on toy data were trained for 50,000 iterations (default setting) and on image data for $200$ epochs. 

\subsection{Toy data}
\begin{figure}[htbp]
\floatconts
  {fig:samples}
  {\caption{Results of density estimation for 2-dimensional toy data (a), and samples of the i-DenseNet trained on CIFAR10 (b).}}
  {%
    \subfigure[Density estimation results after 50,000 iterations of the Residual Flow and i-DenseNet. Trained on 2-dimensional toy data.]{\label{fig:density}%
      \includegraphics[width=0.46\linewidth]{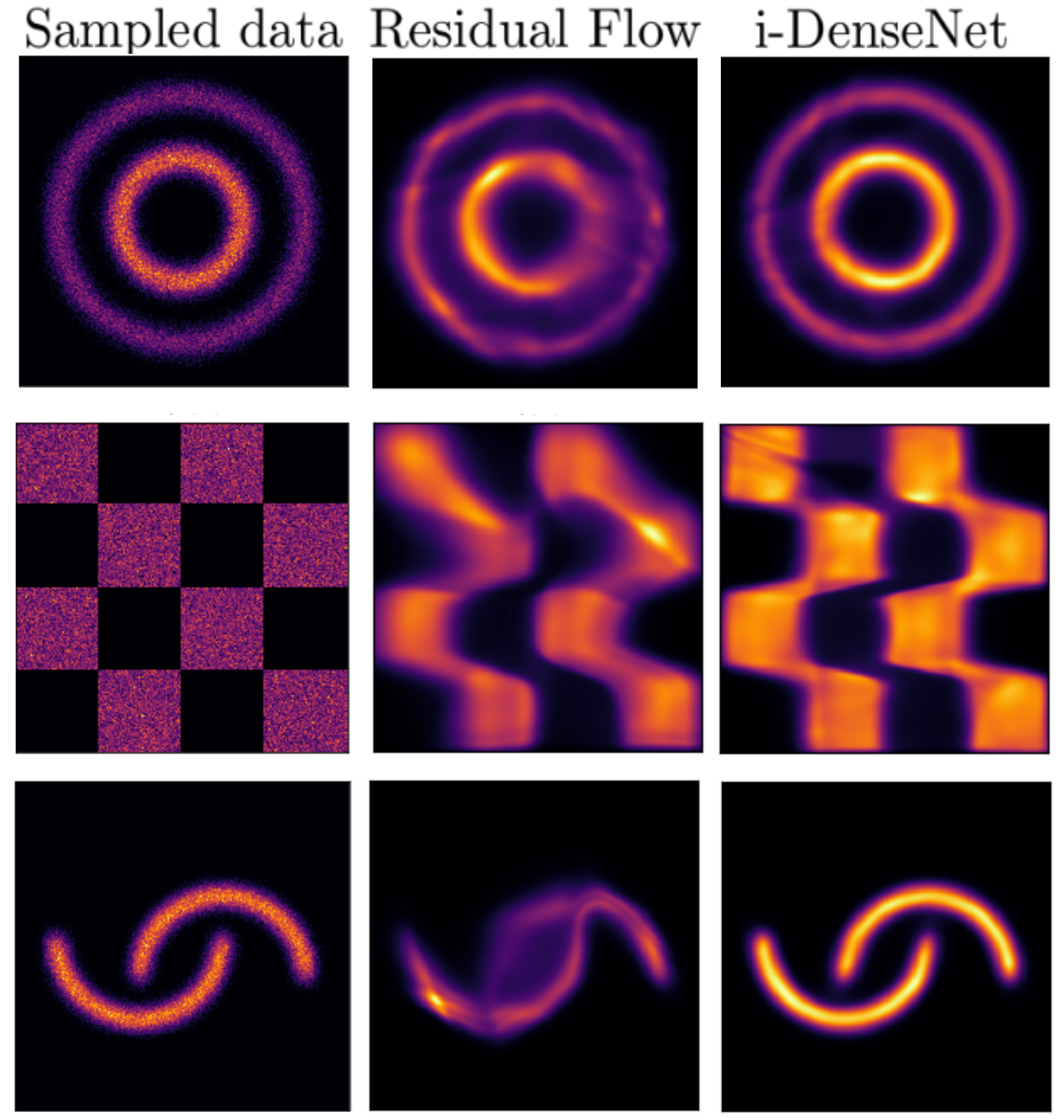}}%
    \qquad
    \subfigure[Samples of i-DenseNet with learnable concatenation.]{\label{fig:samples_dn}%
      \includegraphics[width=0.46\linewidth]{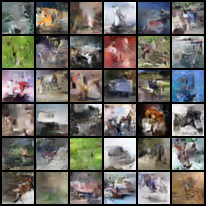}}
  }
\end{figure}

We trained the models on different types of 2-dimensional toy data distributions, namely, two circles, a checkerboard, and two moons. 
The results of the learned density distributions are presented in~\figureref{fig:density}. We observe that Residual Flows are capable to capture high-probability areas. However, they have trouble with learning low probability regions for two circles and moons. i-DenseNets are capable in capturing all regions of the datasets.~\tableref{tab:imgs_bpd}, where the negative log-likelihood for the models are presented, also shows that i-DenseNets with and without learnable concatenation (LC) outperform Residual Flows. The biggest difference in performance is for two moons where i-DenseNets with LC obtain $2.39$ nats compared to $2.60$ nats for Residual Flows. This is consistent with the density estimation plots where i-DenseNets are better in capturing the data distribution than Residual Flows.

\subsection{Image Data}
The results of the models trained on MNIST and CIFAR10 data are presented in~\tableref{tab:imgs_bpd}.
We notice that i-DenseNets outperform Residual Flows in bits per dimension (bpd) on CIFAR10 with $3.41$ bpd without LC and $3.39$ bpd with LC, against $3.42$ bpd for the Residual Flow.~\figureref{fig:samples_dn} presents samples of the i-DenseNet with LC trained on CIFAR10, in~\appendixref{apd:D_appendix} more samples of the models can be found. Additionally, in~\appendixref{apd:C_appendix} a heatmap of the normalized parameters $\hat{\eta}_1$ and $\hat{\eta}_2$ of the learnable concatenation is presented.

During training on MNIST the original Residual Flow suffered from unstable results. This might be due to the coefficient for the spectral normalization, which controls the Lipschitz constraint. In return, this leads to an unstable Jacobian determinant estimation. We adjusted the Lipschitz coefficient for the spectral normalization by setting it to $0.93$ for all models. Additionally, the concatenation in DenseNets is multiplied by $0.98$. Due to slight fluctuations, the results are averaged over the last 5 epochs, which are presented in~\tableref{tab:imgs_bpd}. We observe that i-DeseNets without and with LC outperform the Residual Flow with respectively $1.05$ bpd and $1.04$ bpd against $1.08$ bpd of the Residual Flow. In general we observe that i-DenseNets with LC outperform Residual Flows and i-DenseNets without LC. On two moons, the performance of i-DenseNets with and without LC are tied.

\begin{table}[ht]
    \centering
    \begin{tabular}{c c c c c c c} \toprule
        Model & 2 circles & Checkerboard & 2 moons & MNIST & CIFAR10  \\ \midrule
        Residual Flow & 3.44 & 3.81 & 2.60 & 1.08 & 3.42 \\
        \hline 
        Invertible DenseNet & 3.32 & 3.68 & \textbf{2.39} & 1.05 & 3.41  \\
        \textbf{Invertible DenseNet+LC} & \textbf{3.30} & \textbf{3.66} & \textbf{2.39} & \textbf{1.04} & \textbf{3.39} \\ \bottomrule
    \end{tabular}
    \caption{Negative log-likelihood results on test data in nats (toy data) and bits per dimension (MNIST and CIFAR10). i-DenseNets with and without learnable concatenation are compared with the Residual Flow.}
    \label{tab:imgs_bpd}
\end{table}

%%%%%%%%%%%%%%%%%%%%%%%%%%%%%%%%%%%%%%%%%%%%%%%%%%%%%%%%%%%%%%%%%%%%%%
%                           Conclusion                               %
%%%%%%%%%%%%%%%%%%%%%%%%%%%%%%%%%%%%%%%%%%%%%%%%%%%%%%%%%%%%%%%%%%%%%%
\section{Conclusion} \label{sec:conclusion}
We introduced i-DenseNets, a parameter efficient alternative to Invertible ResNets. Our method enforces invertibility by satisfying the Lipschitz continuity in Dense Layers. In addition, we proposed a version where the concatenation is learned during training, which also indicates which representations are used. We used a smaller architecture under an equal parameter budget, where we demonstrated the performance of i-DenseNets and compared these models to Residual Flows on toy, MNIST, and CIFAR10 data. In conclusion, both i-DenseNets with fixed and learnable concatenation outperformed Residual Flows in negative log-likelihood.

% \acks{Acknowledgements go here.}
\newpage
\bibliography{jmlr-sample}

\newpage
\appendix

%%%%%%%%%%%%%%%%%%%%%%%%%%%%%%%%%%%%%%%%%%%%%%%%%%%%%%%%%%%%%%%%%%%%%%
%                            Derivation                              %
%%%%%%%%%%%%%%%%%%%%%%%%%%%%%%%%%%%%%%%%%%%%%%%%%%%%%%%%%%%%%%%%%%%%%%
\section{Derivation of Lipschitz constant K for the concatenation}
We know that a function $f$ is $\mr{K}$-Lipschitz if for all points $v$ and $w$ the following holds:
\begin{equation} \label{eq:A_lipschitz}
    d_Y(f(v), f(w)) \leq \mr{K} d_X(v, w),
\end{equation}
where $d_Y$ and $d_X$ are distance metrics and $K$ is the Lipschitz constant. 

Consider the case where we assume to have the same distance metric $d_Y = d_X = d$ and where the distance metric is assumed to be chosen as any $p$-norm, where $p \geq 1$, for vectors: $||\delta||_p = \sqrt[\leftroot{-1}\uproot{1}p]{\sum_{i=1}^{len(\delta)} |\delta_i|^p}$. Further, we assume a DenseBlock to be a function $h$ where the output for each data point $v$ and $w$ is expressed as follows:
\begin{equation}
    h_v = 
    \begin{bmatrix}
    f_1(v) \\ f_2(v)
    \end{bmatrix}
    = 
    \begin{bmatrix}
    a_v \\ b_v
    \end{bmatrix}, 
    \quad
    h_w =
    \begin{bmatrix}
    f_1(w) \\ f_2(w)
    \end{bmatrix}
    = 
    \begin{bmatrix}
    a_w \\ b_w
    \end{bmatrix},
\end{equation}
where in this paper for a Dense Layer and for a data point $x$ the function $f_1(x)=x$ and $f_2$ expresses a linear combination of (convolutional) weights with $x$ followed by a non-linearity, for example $\phi(W_1 x)$.
We can re-write \equationref{eq:A_lipschitz} for the DenseNet function as:
\begin{equation} \label{eq:lip_h}
    d(h_v, h_w) \leq \mr{K} d(v, w),
\end{equation}
where $\mr{K}$ is the unknown Lipschitz constant for the entire DenseBlock. However, we can find an analytical form to express a limit on $\mr{K}$. To solve this, we know that the distance between $h_v$ and $h_w$ can be expressed by the $p$-norm as:

\begin{equation} 
\begin{split}
    d(h_v, h_w) &= \sqrt[\leftroot{-1}\uproot{1}p]{\sum_{i=1}^{len(h_v)} |h_{v, i} - h_{w, i}|^p}, \\
\end{split}
\end{equation}
where we can simplify the equation by taking the $p$-th power:

\begin{equation} \label{eq:d_h12}
\begin{split}
    d(h_v, h_w)^p &= \sum_{i=1}^{len(a_v)} |a_{v, i} - a_{w, i}|^p + \sum_{i=1}^{len(b_v)}  |b_{v, i} - b_{w, i}|^p.
\end{split}
\end{equation}
Since we know that the distance of $a$ can be expressed as:
\begin{equation} 
\begin{split}
    d(a_v, a_w) &= \sqrt[\leftroot{-1}\uproot{1}p]{\sum_{i=1}^{len(a_v)} |a_{v, i} - a_{w, i}|^p},
\end{split}
\end{equation}
which is similar for the distance of $b$, re-writing the second term of \equationref{eq:d_h12} in the form of \equationref{eq:lip_h} is assumed to be of form:
\begin{equation} \label{eq:distance_a}
\begin{split}
    d(a_v, a_w)^p &\leq \mr{K}_1^p  d(v, w)^p, \\
\end{split}
\end{equation}
which is similar for $b$, $d(b_v, b_w)^p \leq \mr{K}_2^p  d(v, w)^p$. 
Assuming this, we can find a form of \equationref{eq:lip_h} by substituting with \equationref{eq:d_h12} and \equationref{eq:distance_a}:

\begin{equation} 
\begin{split}
    d(h_v, h_w)^p = \sum_i^{len(h_v)} |h_{v, i} - h_{w, i}|^p &\leq \mr{K}_1^p 
    d(a_v, a_w)^p + \mr{K}_2^p d(b_v, b_w)^p \\
    &= (\mr{K}_1^p +\mr{K}_2^p) d(v, w)^p.
\end{split}
\end{equation}
Now, taking the $p$-th root we have:

\begin{equation} 
    d(h_v, h_w) \leq \sqrt[\leftroot{-1}\uproot{1}p]{(\mr{K}_1^p +\mr{K}_2^p)} d(v, w),
\end{equation}
where we have derived the form of \equationref{eq:lip_h} and where $\mathrm{Lip}(h) = \mr{K}$ is expressed as:

\begin{equation} 
    \mathrm{Lip}(h) = \sqrt[\leftroot{-1}\uproot{1}p]{(\mr{K}_1^p +\mr{K}_2^p)},
\end{equation}
where $\mr{Lip}(f_1) = \mr{K}_1$ and $\mr{Lip}(f_2) = \mr{K}_2$, which are assumed to be known Lipschitz constants. 
 \label{apd:A_appendix}
%%%%%%%%%%%%%%%%%%%%%%%%%%%%%%%%%%%%%%%%%%%%%%%%%%%%%%%%%%%%%%%%%%%%%%
%                            Implementation                          %
%%%%%%%%%%%%%%%%%%%%%%%%%%%%%%%%%%%%%%%%%%%%%%%%%%%%%%%%%%%%%%%%%%%%%%
\section{Implementation}
We used a smaller architecture of Residual Flows~\citep{chen2019residualflows}, with an adjustment of number of blocks per scale set to 4 instead of 16. For training we ensured an equal parameter budget for i-DenseNets. The architecture of i-DenseNets for image data are presented in~\tableref{tab:architecture_dn}. A DenseBlock consist of several Dense Layers.
The last Dense Layer $h_n$ is followed by a $1 \times 1$ convolution to match the output of size $\mathbb{R}^d$ after which a squeezing layer is applied. The final part of the network consist of a Fully Connected (FC) layer with number of blocks set to 4. Before the concatenation in the FC layer, a Linear layer of input $\mathbb{R}^d$ to output dimension $64$ is applied, followed by the Dense Layer with DenseNet growth $32$ and activation LipSwish. The DenseNet depth is set to $3$. The final part consist of a Linear layer to match the output of size $\mathbb{R}^d$.

\begin{table}[htbp]
\begin{tabular}{cccccc}
\toprule
\multirow{2}{*}{\begin{tabular}[c]{@{}c@{}}Nr. \\ of scales\end{tabular}} & \multirow{2}{*}{\begin{tabular}[c]{@{}c@{}}Nr. of blocks \\ per scale\end{tabular}} & \multirow{2}{*}{\begin{tabular}[c]{@{}c@{}}DenseNet\\ Depth\end{tabular}} & \multirow{2}{*}{\begin{tabular}[c]{@{}c@{}}DenseNet\\ Growth\end{tabular}} &
\multirow{2}{*}{Dense Layer} & 
\multirow{2}{*}{Output} \\
&   &   &   &   & \\ \midrule
$3$  & $4$  & $3$ 
& 
\multicolumn{1}{l}{\begin{tabular}[c]{@{}l@{}}108 (MNIST)\\ 124 (CIFAR10)\end{tabular}} 
& 
$\begin{bmatrix}
3 \times 3 \quad \mr{conv} \\
\mr{LipSwish} \\
\text{concat}
\end{bmatrix}$
& 
$\begin{bmatrix}
1 \times 1 \quad \mr{conv}
\end{bmatrix}$
\\ \bottomrule
\end{tabular}
\caption{The architecture for function $g$ for image data.}
\label{tab:architecture_dn}
\end{table}

\paragraph{Toy data} We used 10 scale blocks for all models. Furthermore, we used default settings of Residual Flows. For i-DenseNets, we choose a DenseNet -depth and -growth of, respectively, 4 and 90 with $504$K parameters and Residual Flows utilize $501$K parameters. 

\paragraph{MNIST} All models used $3$ scales where the number of blocks per scale is set to 4. Due to instability of Residual Flows, we set our coefficient that controls the Lipschitz constraint from $0.98$ to $0.93$. Furthermore, default settings of Residual Flows are used. For i-DenseNets, we used a coefficient controlling the Lipschitz of the concatenated blocks set to $0.98$. i-DenseNets use a DenseNet -depth and -growth of, respectively, 3 and 108 with $5.0$M parameters and Residual Flows utilize $5.0$M parameters.

\paragraph{CIFAR10} All models used $3$ scales where the number of blocks per scale is set to 4. Furthermore, default settings of Residual Flows are used. i-DenseNets use a DenseNet -depth and -growth of, respectively, 3 and 124 with $8.7$M parameters and Residual Flows utilize the $8.7$M parameters.
 \label{apd:B_appendix}
%%%%%%%%%%%%%%%%%%%%%%%%%%%%%%%%%%%%%%%%%%%%%%%%%%%%%%%%%%%%%%%%%%%%%%
%                Visualization of learnable concat                   %
%%%%%%%%%%%%%%%%%%%%%%%%%%%%%%%%%%%%%%%%%%%%%%%%%%%%%%%%%%%%%%%%%%%%%%
\section{Visualization of learnable concatenation}
\begin{figure}[htbp]
\floatconts
  {fig:heatmaps}
  {\caption{Heatmaps of the normalized $\eta_1$ and $\eta_2$ after training for 200 epochs on CIFAR10.}}
  {%
    \subfigure[Heatmap of $\hat{\eta}_1$][c]{\label{fig:eta1}%
      \includegraphics[width=.9\linewidth]{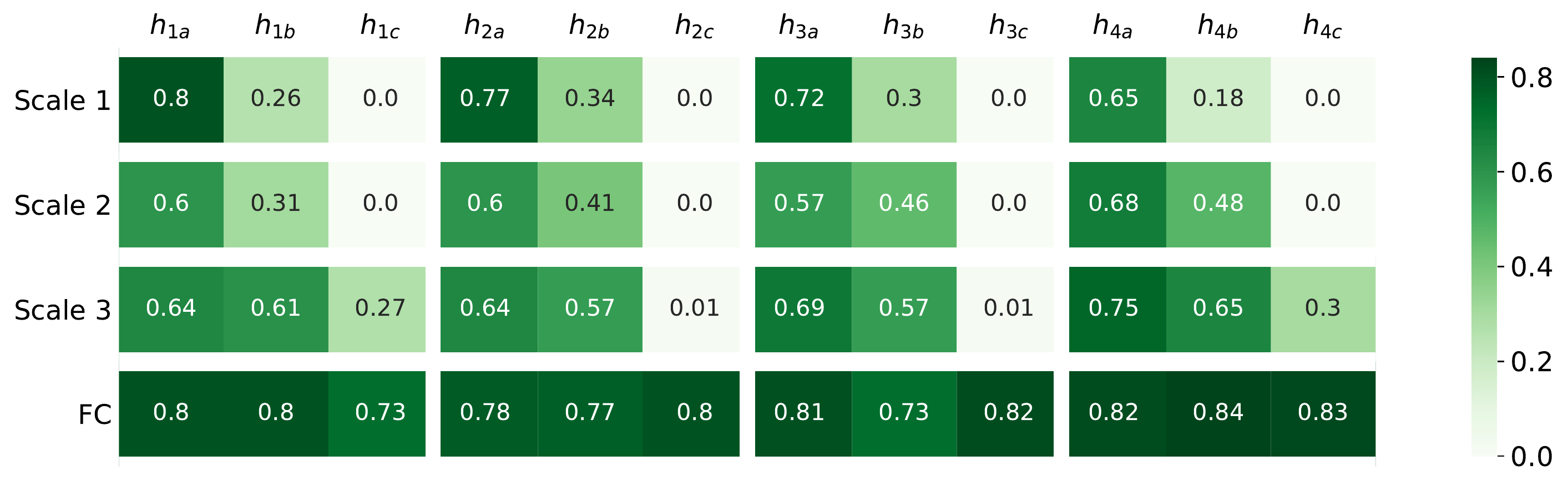}}%
    \par \medskip
    \subfigure[Heatmap of $\hat{\eta}_2$.][c]{\label{fig:eta2}%
      \includegraphics[width=.9\linewidth]{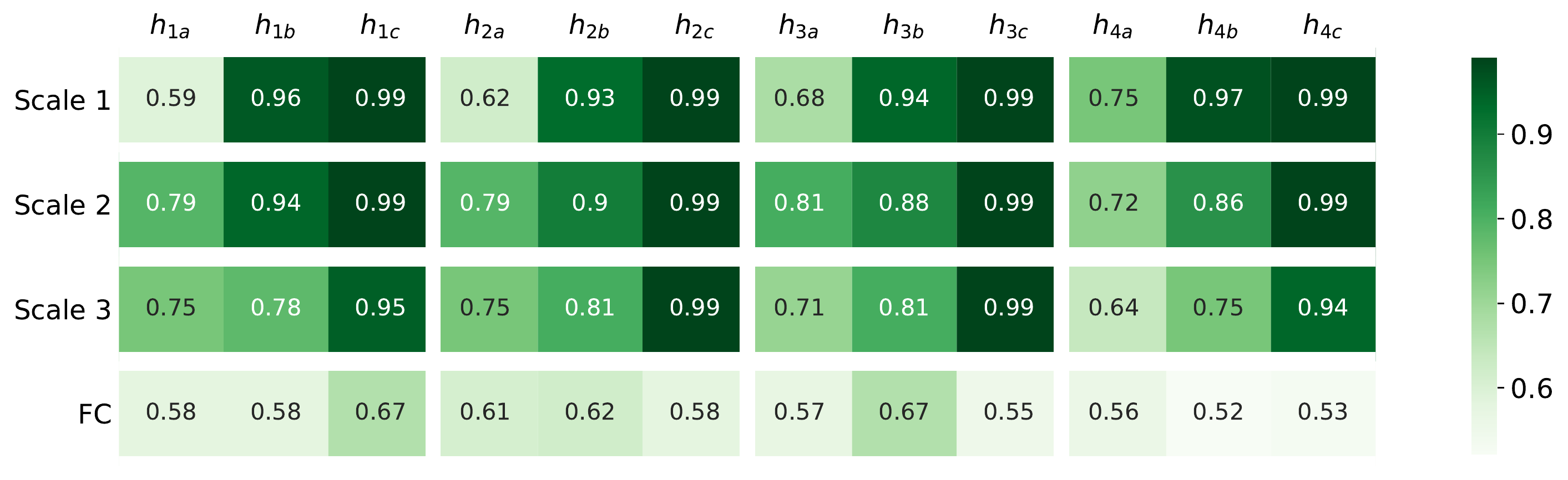}}
  }%
\end{figure}
\noindent \figureref{fig:heatmaps} shows the heatmap for \subfigref{fig:eta1} the normalized parameter $\hat{\eta}_1$ and \subfigref{fig:eta2} normalized parameter $\hat{\eta}_2$ after 200 epochs, trained on CIFAR10. 
Every scale level 1, 2 and 3 contain 4 DenseBlocks, that each contain 3 Dense Layers with convolutional layers. 
The final level FC indicates that fully connected layers are used. The letters `a', `b', and `c' index the Dense Layers per block. Remarkably, all scale levels for the last layers $h_{ic}$ give little importance to the input signal. The input signals for these layers are in most cases multiplied with $\hat{\eta}_1$ (close to) zero, while the transformed signal uses almost all the information when multiplied with $\hat{\eta}_2$ which is close to one. This indicates that the transformed signal is of more importance for the network than the input signal. For the fully connected part, this difference is not that pronounced.
 \label{apd:C_appendix}
%%%%%%%%%%%%%%%%%%%%%%%%%%%%%%%%%%%%%%%%%%%%%%%%%%%%%%%%%%%%%%%%%%%%%%
%                               Samples                              %
%%%%%%%%%%%%%%%%%%%%%%%%%%%%%%%%%%%%%%%%%%%%%%%%%%%%%%%%%%%%%%%%%%%%%%
\section{Model samples}
\begin{figure}[htbp]
\floatconts
  {fig:mnist}
  {\caption{Results on MNIST.}}
  {%
  \subfigure[Real images.]{\label{fig:mnist_real}%
      \includegraphics[width=0.46\linewidth]{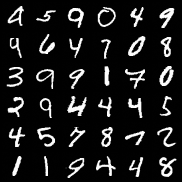}}
    \qquad
    \subfigure[Samples of the Residual Flow.]{\label{fig:mnist_res}%
      \includegraphics[width=0.46\linewidth]{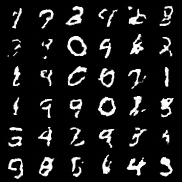}}
  \par \medskip
      \subfigure[Samples of i-DenseNet without LC.]{\label{fig:mnist_dn}%
      \includegraphics[width=0.46\linewidth]{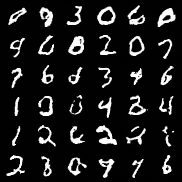}}%
    \qquad
\subfigure[Samples of i-DenseNet with LC.]{\label{fig:mnist_dn_lc}%
      \includegraphics[width=0.46\linewidth]{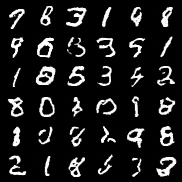}}%
  }
\end{figure}

This appendix contains real images and samples of the models trained on MNIST (\figureref{fig:mnist}). \subfigref{fig:mnist_real} shows real images of MNIST, \subfigref{fig:mnist_res} shows samples of the Residual Flow trained on MNIST, as well as samples of \subfigref{fig:mnist_dn} i-DenseNet without LC and \subfigref{fig:mnist_dn_lc} i-DenseNet with LC. 

\begin{figure}[htbp]
\floatconts
  {fig:cifar10}
  {\caption{Results on CIFAR10}}
  {%
  \subfigure[Real images.]{\label{fig:cifar10_real}%
      \includegraphics[width=0.46\linewidth]{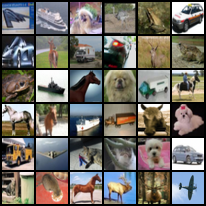}}
    \qquad
    \subfigure[Samples of the Residual Flow.]{\label{fig:cifar10_res}%
      \includegraphics[width=0.46\linewidth]{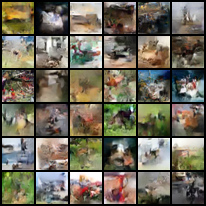}}
  \par \medskip
      \subfigure[Samples of i-DenseNet without LC.]{\label{fig:cifar10_dn}%
      \includegraphics[width=0.46\linewidth]{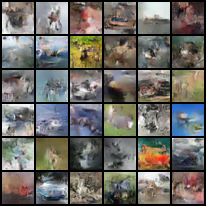}}%
    \qquad
    \subfigure[Samples of i-DenseNet with LC.]{\label{fig:cifar10_dn_lc}%
      \includegraphics[width=0.46\linewidth]{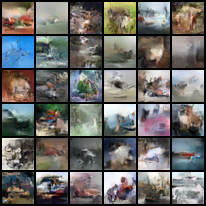}}%
  }
\end{figure}

\figureref{fig:cifar10} contains real images and samples of the models trained on images of CIFAR10. \subfigref{fig:cifar10_real} shows real images of CIFAR10, \subfigref{fig:cifar10_res} shows samples of the Residual Flow trained on CIFAR10, as well as samples of \subfigref{fig:cifar10_dn} i-DenseNet without LC and \subfigref{fig:cifar10_dn_lc} i-DenseNet with LC.

 \label{apd:D_appendix}

\end{document}